# Mobile APP User Attribute Prediction by Heterogeneous Information Network Modeling

Hekai ZHANG[1,2], Jibing GONG[1,2,5], Zhiyong TENG[1,2], Dan WANG[1,2]
Hongfei WANG[3], Linfeng DU[4] and ZAKIRUL ALAM BHUIYAN[6]

hekai_zhang@163.com  gongjibing@163.com  tengzhiyongdgqb@gmail.com
wangdanysu8100@163.com  gabrielle_w@buaa.edu.cn  fred9918@buaa.edu.cn
ZAKIRULALAM@GMAIL.COM

[1] School of Information Science and Engineering, Yanshan University,
Qinhuangdao 066004, China;
[2] The Key Lab for Computer Virtual Technology and System Integration,
Yanshan University, Qinhuangdao 066004, China
[3] School of Computer Science and Engineering, Beihang University, Beijing 100191, China
[4] Shenyuan Honors College, Beihang University, Beijing 100191, China
[5] State Key Lab of Mathematical Engineering and Advanced Computing, Wuxi 214000, China
[6] Department of Computer and Information Sciences, Fordham University

**Abstract.** User-based attribute information, such as age and gender, is usually considered as user privacy information. It is difficult for enterprises to obtain user-based privacy attribute information. However, user-based privacy attribute information has a wide range of applications in personalized services, user behavior analysis and other aspects. Although many scholars have made achievements in user attribute prediction and other related fields, there are still two main problems that impede further improvement on the accuracy of classification: (1) Traditional machine learning classification merely takes each object as a single individual, ignoring the relationship between them; (2) At present, the popular Heterogeneous Path-Mine Information Network only considers whether the user has a relationship with the attributes of other nodes, rather than the degree of correlation of the attributes. It employs a linear regression model to fit the weight of meta-path, which is highly sensitive to outliers. To solve the above two problems, this paper advances the HetPathMine model and puts forward TPathMine model. With applying the number of clicks of attributes under each node to express the user's emotional preference information, optimizations of the solution of meta-path weight are also presented. Based on meta-path in heterogeneous information networks, the new model integrates all relationships among objects into isomorphic relationships of classified objects. Matrix is used to realize the knowledge dissemination of category knowledge among isomorphic objects. The experimental results show that: (1) the prediction of user attributes based on heterogeneous information networks can achieve higher accuracy than traditional machine learning classification methods; (2) TPath-

Mine model based on the number of clicks is more accurate in classifying users of different age groups, and the weight of each meta-path is consistent with human intuition or the real world situation.

# 1 Introduction

For mobile phone device manufacturers, it is challenging to obtain the current user's population privacy attribute information. However, user's private attribute information is crucial for enterprises, given their important role played in optimizing services and so on. In recent years, with the widespread popularity of the Internet, researchers have used various algorithms to study user attributes based on users' online behavior, especially browsing habits. Those proposed approaches would not violet general data protection regulations[21-22]. For example, Chen et al. [1] used mixed classification/regression model to predict users' age; Wang et al. [2] used a support vector model to predict the age attributes of users that showing good results. Information network is an effective way to express the relationship between data [8-9]. Thus, researchers pay more and more attention to it [18-20]. In particular, the classification and clustering of heterogeneous information networks have been one of the research hot-spots [16-17]. For example, in Luo et al. [14], a HetPathMine algorithm is proposed, which uses information network association to propagate classification knowledge from labeled nodes to unlabeled nodes, and achieves classification algorithm. This algorithm is currently the most widely used classification method for heterogeneous information networks.

However, two main limitations to these methods are remained: (a) they ignore the heterogeneous information in the data. The traditional machine learning classification method only treats each object as a single individual, ignoring the connection between the object and the object, so that the rich semantic information hidden in the relationship between different entities in the data is unexploited. (b) The more popular HetPathMine heterogeneous information network classification model objects can only reflect if there is a relationship (1 means that there is a relationship, 0 means no relationship), but is lack of the degree of the relationship, thus leading to the qualified use of user's behavior data to have a more accurate classification. For example, even the user downloads an app, it can be uninstalled as a result of not interested. This is an abnormal behavior, but it can still be captured by the heterogeneous information network. Moreover, it uses the least squares method in linear regression to solve the meta-path weight, making the model extremely sensitive to abnormal points.

These limitations prompt us to design a model that can learn more comprehensive user behavior information, with a superior classification capability. Therefore, we propose a method that uses the number of clicks to express the user's emotional preferences, applying to heterogeneous information networks, and thus improving the meta-path weight fitting manner. In order to solve the above two limitations, in this paper, we propose a new model based on the original HetPathMine model, called

TPathMine, to predict the user attributes of Huawei Account. We use the number of clicks to record the relationship matrix of each object, instead of just 0 and 1 to indicate whether there is a click relationship. TPathMine model learns annotated nodes and achieves the classification of heterogeneous objects including users' emotional preferences through knowledge dissemination.

## 2      Related Work

A large number of tasks such as user classification are effectively solved by machine learning and other classification methods. Weber et al. [5] based on Yahoo search query log and user profile, analyzed the relationship between user characteristics and query content, and found that people with similar user characteristics were more likely to search for similar items. Xu et al. [6] successfully classified the age and gender of users based on Bayesian theory according to the historical records of users browsing the website. Class prediction; Bi et al. [7] used the search habits of users with Bing, combined with Facebook data to infer the age and gender of users, and also achieved good classification results.

Information network is an effective way to express data. Up to now, many heterogeneous information network classification algorithms have been proposed in academia. For example, Zhou [13] and others proposed LLGC algorithm, which uses network structure to transmit labeled data to unlabeled data. LLGC algorithm is a very famous algorithm in network mining. Liu [4] proposes another network semi-supervised classification algorithm, which considers all neighbor nodes when labeling unclassified nodes. The above two algorithms are proposed on isomorphic networks. Ming et al. [15] proposed a semi-supervised classification algorithm in heterogeneous information networks: GNetMine. However, GNetMine assumes that all nodes in heterogeneous information networks have the same classification method. However, in heterogeneous information networks, there are many types of nodes and edges, and there are different classification standards for nodes and edges. In Luo et al. [14], a HetPathMine algorithm is proposed, which integrates all relationships among objects into the isomorphic matrix of classified objects based on meta-paths in heterogeneous information networks and achieves knowledge dissemination among isomorphic objects.

Since our algorithm is based on the enhancement of HetPathMine, we focus on the HetPathMine algorithm. The algorithm builds multiple relational meta-paths between isomorphic objects based on multiple knowledge propagation paths among heterogeneous objects. The multiple meta-paths among these isomorphic objects integrate all relationships among objects in the heterogeneous information network. Then the weights of each meta-path relationship matrix are calculated by logical regression method. In line with the weights, the multiple meta-paths are fused together to form the isomorphic object relationship matrix. Then, according to the transmission of isomorphic knowledge, the isomorphic object relationship matrix is formed. Sowing implements classification [3]. Luo's method allows different types of nodes to have different classification criteria, and determines the weight of each path to the classifi-cation results through logical regression, rather than handling them equally. On the other hand, the algorithm has inadequacies: (1) it only indicates whether there is a relationship between objects, but not shows the degree of the relationship, and it is easy to capture some abnormal data; (2) With the least square method in linear regression to fit the weight of meta-path, it is sensitive to abnormal information, which leads to the inability to base on users. Behavior is classified more accurately.

## 3 Data Analysis

### 3.1 Data Analysis

The method we propose is to evaluate the dataset obtained from "Huawei Account User Attributes", as one of the largest mobile phone user platforms in China. Detailed are shown in Table 1.

**Table 1.** An Overview of the Data Set of
"Population Attributes of Huawei Account Users"

| Nodes | Count | Links | Count |
|---|---|---|---|
| app | 167622 | app-user | 33154654 |
|  |  | app-type | 188864 |
| user | 2110000 | user-app | 33154654 |
|  |  | user-type | 53418679 |
| type | 40 | type-app | 188864 |
|  |  | type-user | 53418679 |
| Total | 2177662 | Total | 173524394 |

This paper regards the prediction problem of user age as a classification problem, and divides the user's age stage into six categories, of which 1 represents less than or equal to 18 years old, 2 represents 19-23 years old, 3 represents 24-34 years old, 4 represents 35-44 years old, 5 represents 45-54 years old, and 6 represents greater than or equal to 55 years old. The application type consists of 40 types including "book reading", "tourism accommodation", "child", "business", etc. Each app application belongs to at least 1 application type, and at most 8 application types. In the 33,154,654 user-app relationship, each user clicked on an average of 12 app applications. People of different ages tend to use different types of applications and have a great dissimilarity in the number of uses for each category. Different application types have different needs, which also verifies the importance of user attribute prediction.

### 3.2 Meta Path Analysis

There are three types of objects in this network: users, app applications and application types. The link between user and app application consists of the relationship be-tween "click" and "clicked". The relationship between app application and application type is the relationship between "belong" and "belonged to" and the relationship between user and application type is the relationship between "find" and "be found". Currently, we analyze and finally obtain four types of meta-path shown in Figure 1.

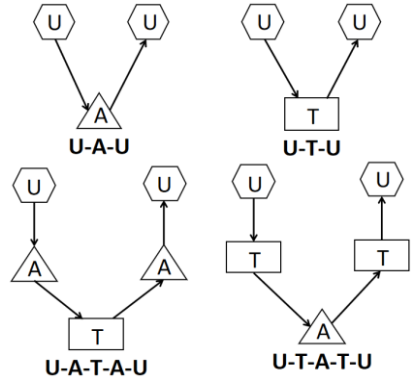

**Fig 1.** Analyzed meta-paths

At present, we analyze and finally get the four types of meta-paths shown as follows. Semantic meanings are described as follows:

- U-A-U: Two users have clicked on the same app application.
- U-T-U: Two users have looked up the same application category.
- U-A-T-A-U: Apps clicked by two users belong to the same category.
- U-T-A-T-U: The different application categories that two users find contain the same app application.

## 4 Methodology

### 4.1 An Overview of TPathMine

Figure 2. shows an overview of the network architecture of our proposed model. This framework mainly consists of three submodules. (a)Meta-path sampling in HIN. In this part, we first construct HIN among users, app and type, and utilize random walk to sample meta-paths on this network for each given user behavior. (b)Path-attention calculation. In order to understand which meta-path is affected in the process of classifying users, attention mechanism is adopted here, and the weight of each meta-path relationship matrix is calculated by SVR method. (c)Age classification of person. Then, according to the weight, Multiple meta-paths are fused together to form isomorphic object relationship matrix, and then classified according to the dissemination of isomorphic knowledge.

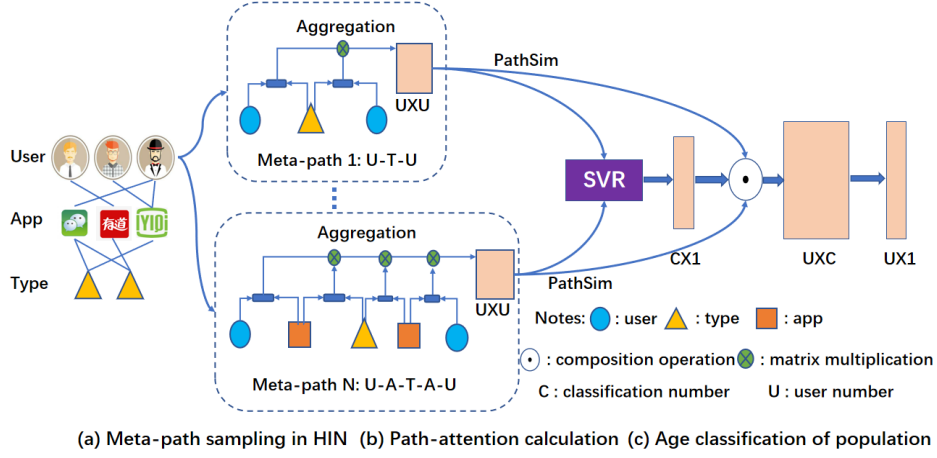

(a) Meta-path sampling in HIN  (b) Path-attention calculation  (c) Age classification of population

**Fig 2.** The overall architecture of the proposed model.

### 4.2   The TPathMine Model

Given a set of meta-paths, expressed as $P = \{p_1, p_2, ..., p_d\}$, where d is the number of meta-paths used in our algorithm. Then, the weight of each meta-path is expressed as $\beta = \{\beta_1, \beta_2, ..., \beta_d\}$. We use support vector regression (SVR) algorithm[10] to calculate $\beta$. The optimization function is as follows:

$$\begin{cases} \min \quad \frac{1}{2}\|\beta\|^2 \\ s.t. \quad |R(x_i, x_j) - (\beta S(x_i, x_j) + b)| \leq \varepsilon \quad \forall i, j \end{cases} \quad (1)$$

where $S(x_i, x_j)$ is a vector, defined as the PathSim metric of $x_i$ and $x_j$ nodes under different meta-paths. We set $S(x_i, x_j) = (s_1(x_i, x_j), s_2(x_i, x_j), \cdots, s_d(x_i, x_j))$ as a training sample. R is the relationship matrix between different nodes, which can be calculated as follows:

$$R(x_i, x_j) = \begin{cases} n, & \text{number of connections between } x_i \text{ and } x_j \\ 0, & \text{others} \end{cases} \quad (2)$$

$\varepsilon$ is the error adjustment factor used to control the accuracy of the fit. Through SVR, we can calculate a set of optimal, and $\beta$ assigns reasonable weights to different meta-paths.

Du et al. [12] proposed a semi-supervised classification algorithm on isomorphic networks. We extended this algorithm to heterogeneous networks. The extended model is as follows:

$$\xi(F) = \frac{1}{2}(\sum_{k=1}^{d}\sum_{i=0}^{n}\sum_{j=0}^{n}\beta_k W_{ij}^{k} \| \frac{F_i}{\sqrt{D_{ii}^{k}}} - \frac{F_j}{\sqrt{D_{jj}^{k}}} \|^2) + \lambda \sum_{i=0}^{n}\sum_{i=0}^{n} \| F_i - Y_i \|^2 \quad (3)$$

In the above, n is the number of objects within the target type $X_i$, $W^k$ is the similarity matrix for the target type reduced by the $k-th$ meta-path. $F$ is a $n*p$ matrix, where p is the number of class, and $F(i,j)$ denotes the probability of $i-th$ object belonging to the $j-th$ class. $Y$ is also a $n*p$ matrix which denotes the pre-labeled information in the network, $D^k$ is a diagonal matrix with its $(i,i)-th$ element equal to the sum of the $i-th$ row of $W^k$. $\beta = \{\beta_1, \beta_2,..., \beta_d\}$ is calculated by Eq. (1). Then, we can get result as follow:

$$\frac{\partial \xi}{\partial F} = F^* - F^*(\sum_{k=1}^{m} \beta_k S^k) + \mu(F^* - Y) \quad (4)$$

$$F^* = \beta(I - aS_{com})^{-1} Y \quad (5)$$

Where $S = D^{k^{-1/2}} W^k D^{k^{-1/2}}$, $a = \frac{1}{1+\mu}$, $S_{com} = \sum_{k=1}^{m} \beta_k S^k$.

After getting $F$, We can get the category labels of all objects by following formula:

$$Lable(x_i) = \max\{F(i,1), F(i,2),..., F(i,n)\} \quad (6)$$

It is pointed out that, the output of TPathMine not only contains the classification result but also the different weight $\beta = \{\beta_1, \beta_2,..., \beta_d\}$ for the selected meta-path, $\beta$ can be used in many data mining tasks[17,18].

## 5 Experimental Results

### 5.1 Experiment Details

In reality, a large number of users with unknown attributes are often hidden, and we now have very few user data. In order to solve the problem of lack of labels in practical applications, we randomly select a% = [10%, 20%, ..., 50%] percent of users use their label information as a training set in classification tasks. Seed fraction determines the percentage of marker information. For each given seed part of the data set, our final accuracy results are 5 different choices based on the seed, and the average of the calculated results is taken as the final accuracy value. The classification results are shown in Table 2. In our experiment, we will transform the classification model to set parameter $\lambda = 2$.

Overall, TPathMine performed best on all data sets with different seed. It is better than Knn, Decision Tree and Svm, which are 16.68 percentage points higher than the traditional machine learning classification algorithm. It is more effective than the most advanced HetPathMine classification algorithm. The results shown in Table 3 also show that TPathMine maintains high accuracy in each data set and can obtain more meaningful results than traditional machine learning classification by mining heterogeneous networks.

**Table 2.** Experimental data results

|      | Knn   | Decision tree | Svm   | HetPathMine | TPathMine |
|------|-------|---------------|-------|-------------|-----------|
| **10%** | 35.76 | 38.95 | 42.15 | 62.15 | 68.52 |
| **20%** | 39.21 | 42.13 | 49.57 | 64.68 | 74.76 |
| **30%** | 44.57 | 44.26 | 53.21 | 69.56 | 75.67 |
| **40%** | 50.21 | 49.58 | 60.81 | 72.7  | 78.05 |
| **50%** | 55.76 | 56.73 | 66.92 | 75.21 | 83.6  |

### 5.2 Parameters Analysis

Here, we study the sensitivity of parameters. In order to verify the validity of transformation classification in HIN, we analyze the performance of TPathMine model when $\lambda$ changes. The parameter $\lambda$ is set to 1, 2, 4, 6, 8 in turn. The experimental results are shown in Figure 3(a). We can see that under the same $\lambda$, the more tags the data sets have, the better the classification effect. In different tagged data sets, when $\lambda = 2$, we can actually make our TPathMine model get better performance in age classification, otherwise the model will get lower performance.

$\varepsilon$ is the error adjusting factor in SVR, which is used to control the accuracy of fitting. A lot of experiments have been carried out to verify the validity of parameter $\varepsilon$. Firstly, we set $\varepsilon$ to 0.1 to 0.5 and show their experimental results in Figure 3(b). We can see that when $\varepsilon = 0.2$, the performance of the model is slightly improved. Excessive setting of $\varepsilon$ can easily lead to over-fitting of the model, and too low cannot effectively identify outliers.

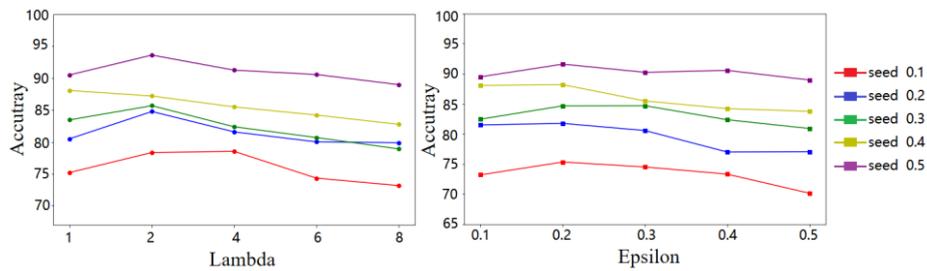

**Fig 3.** (a)Accuracies under Different $\lambda$ Parameters;(b)Accuracies under Different $\varepsilon$ Parameters

For illustration purposes, the weight of each meta-path is mapped to the range [0,1], the weight of each meta-path is shown in Table 3. The meta-path "User-App-User" has the lowest weight in the TPathMine model, which is consistent with the real-world scenario: since each app application has multiple application categories, users may click on the app application just to use one of these application categories, thus expanding the use of the crowd. The meta-path "User-App-Type-App-User" always has the greatest weight compared to the other three meta-paths, which is the same as human intuition: people of the same age group have roughly the same preference for the application category, which is consistent with our perception of real life.

Table 3. Meta-path weight comparison

| ID | Meta-path | Weight |
|---|---|---|
| 1 | User-App-User | 0.05~0.15 |
| 2 | User-Type-User | 0.25~0.35 |
| 3 | User-App-Type-App-User | 0.45~0.55 |
| 4 | User-Type-App-Type-User | 0.15~0.25 |

## 6  Conclusions

In order to incorporate user's clicks into the model and make better use of user's preference information, this paper proposes a classification model TPathMine, which employs the number of clicks to represent user's emotional information. The model takes into account the relationships between all structural objects, including isomorphic relationships (such as users and users) and heterogeneous relationships (such as user and application types). In order to understand which meta-path play a role in the classification process, we assign a weight to each meta-path and use a better SVM algorithm to fit them. The experimental results show that (1) user attribute prediction based on heterogeneous information network can obtain higher accuracy than traditional machine learning classification method; (2) TPathMine model based on user clicks is more accurate for users of different age groups. Also, the weight obtained of each meta-path is consistent with human intuition or real-world conditions.